\documentclass{article}

\usepackage{microtype}
\usepackage{graphicx}
\usepackage{subfigure}
\usepackage{booktabs} 
\usepackage{amsmath}
\usepackage{enumitem}
\usepackage{comment}

\usepackage{hyperref}

\usepackage[accepted]{icml2021}

\usepackage{amssymb}
\usepackage[scaled=.7]{beramono} 
\def\ie{{i.e.},~}

\usepackage{microtype}
\usepackage{threeparttable}

\usepackage{color, colortbl}
\definecolor{Gray}{gray}{0.9}
\usepackage{multirow}
\usepackage{xspace}
\usepackage{caption}
\usepackage{algorithm}
\usepackage{algorithmic}
\usepackage{float}
\usepackage{bm}

\newcommand{\shortsectionBf}[1]{\vspace{0pt}
\noindent {\bf #1}}
\newcommand{\sys}{{\texttt{PersFL}\xspace}}

\usepackage{pifont}
\newcommand{\xmark}{\ding{55}}%

\usepackage{tikz}

\icmltitlerunning{New Metrics to Evaluate the Performance and Fairness of Personalized Federated Learning}

\begin{document}

\twocolumn[
\icmltitle{New Metrics to Evaluate the Performance and Fairness of Personalized Federated Learning}




\begin{icmlauthorlist}
\icmlauthor{Siddharth Divi}{pur}
\icmlauthor{Yi-Shan Lin}{pur}
\icmlauthor{Habiba Farrukh}{pur}
\icmlauthor{Z.Berkay Celik}{pur}
\end{icmlauthorlist}

\icmlaffiliation{pur}{Department of Computer Science, Purdue University, USA}

\icmlcorrespondingauthor{Siddharth Divi}{siddhudivi@gmail.com}

\icmlkeywords{Machine Learning, ICML}

\vskip 0.3in
]



\printAffiliationsAndNotice{
} 

\begin{abstract}
In Federated Learning (FL), the clients learn a single global model (\texttt{FedAvg}) through a central aggregator. In this setting, the non-IID distribution of the data across clients restricts the global FL model from delivering good performance on the local data of each client. Personalized FL aims to address this problem by finding a personalized model for each client.
Recent works widely report the average personalized model accuracy on a particular data split of a dataset to evaluate the effectiveness of their methods.
However, considering the multitude of personalization approaches proposed, it is critical to study the per-user personalized accuracy and the accuracy improvements among users with an equitable notion of fairness. 
To address these issues, we present a set of performance and fairness metrics intending to assess the quality of personalized FL methods.
We apply these metrics to four recently proposed personalized FL methods, \sys, \texttt{FedPer}, \texttt{pFedMe}, and \texttt{Per-FedAvg}, on three different data splits of the CIFAR-10 dataset.
Our evaluations show that the personalized model with the highest average accuracy across users may not necessarily be the fairest.
Our code is available at \url{https://tinyurl.com/1hp9ywfa} for public use.

%
%
%
%
%
%
\end{abstract}

\section{Introduction}
\label{sec:introduction}


Federated Learning (FL) is a distributed collaborative learning paradigm that does not require centralized data storage in a single location. Instead, a joint global predictor is learned by a network of participating users~\cite{fedAvg}.
FL is useful when the clients have sensitive data that they cannot share with the participating entities due to privacy concerns. 
Yet, despite its widespread applications, FL faces different challenges, such as expensive communication, systems heterogeneity, statistical heterogeneity, and privacy concerns~\cite{flSurvey_1}. Among these, statistical heterogeneity has recently gained attention. %

%
%

\shortsectionBf{Statistical Heterogeneity Problem.}
Statistical heterogeneity means that the clients have unbalanced and non-identical and independently distributed (non-IID) data.
This causes the global model (\texttt{FedAvg}) trained on non-IID data of clients not to generalize well on the clients' local data.
Consider the task of predicting the next word on a smartphone, which enables users to express themselves faster.
In such settings, a global model learned collaboratively fails to give personalized suggestions to each user as each user has a unique way of expressing themselves in applications, such as text messaging and writing e-mails.
On the other hand, learning a local model without user collaboration might yield large model error due to the lack of data.
%

\shortsectionBf{Personalized Federated Learning.} 
The personalized learning methods aim to address this problem by learning a personalized model for each client that benefits from the data of the other clients while at the same time overcoming the problem of statistical heterogeneity.
These methods learn a personalized model by extending meta-learning, local fine-tuning,  multi-task learning,  model regularization, contextualization, and model interpolation (See Section~\ref{sec:related_works}).

These efforts often solely report the average accuracy of personalized models across all users to measure their effectiveness.
Yet, the average accuracy does not capture the notion of per-user personalization, as per-user performance is aggregated into a single (averaged) accuracy metric.
Furthermore, they do not measure the fairness of the personalized models from an equitable notion, the concept that the users get similar improvements~\cite{qFFL}.  
The lack of fairness analysis makes it difficult to compare how different personalized models perform on each user.
Lastly, these works employ different data-split strategies among users though they often use standard datasets, such as MNIST and CIFAR-100. 
Overall, these issues make it difficult for a uniform comparison of the effectiveness of each method.

\shortsectionBf{Contributions.}
We present a set of metrics in two groups, five metrics for \texttt{performance} and four metrics for \texttt{fairness}, to assess the quality of personalized FL methods, supporting the existing evaluation metrics.
Performance metrics express how well the personalized model performs over each user's local and global model accuracy.
On the other hand, fairness metrics express an equitable notion that quantifies whether the personalized models provide an equal improvement upon each user's local and global models.
The metrics allow for quantitatively contrasting the trade-off between fairness and per-user accuracy of the personalized models under different datasets and data splits.

To motivate the need for new metrics for personalized FL, we have surveyed $12$ recent works with the goal of studying their datasets, data splitting strategies, and reported evaluation metrics.
We found that these works often use a different data splitting strategy on different datasets and solely report the average accuracy improvement of the personalized model over the global model. 
We evaluate the proposed performance and fairness metrics on four recent personalized FL methods across three different data splits on the CIFAR-10 dataset.
Our evaluation results show that the personalized model with the highest average accuracy across users need not necessarily be the fairest.


\section{Related Work}
\label{sec:related_works}
There exist several recent methods proposed for personalization in FL. 
In local fine-tuning, each user adapts a copy of the global \texttt{FedAvg} model to their local data distribution through gradient-based meta-learning, transfer learning~\cite{transferLearning_survey} and domain-adaptation~\cite{domainAdaptation_base}.
For instance, persFL~\cite{persFL} combines the idea of generalized distillation with optimal teacher models for each user to learn more personalized models.
Per-FedAvg~\cite{metaFL_2} uses Model-Agnostic Meta-Learning~\cite{maml} to learn a common initialization point for each user during training, which is then subsequently adapted to each user's local data distribution. 
FedPer~\cite{fedPer} views the network as a combination of base and personalization layers where the base layers are learned collaboratively, and the personalized layers are specific to each user. 

Previous works have also explored contextualization, which aims at learning a model under different contexts.
This problem is studied in the next character recognition task~\cite{hard_etal}, which needs access to features about the context during the training phase.
Local-Global Federated Averaging (LG-FedAvg)~\cite{lgFedAvg} learns compact local representations on each local and the global model across all users, \ie an ensemble of local and global models.
%

Models can also be personalized to each user by regularizing the differences between the global and local models. 
pFedMe~\cite{pFedMe} uses Moreau envelopes~\cite{moreauEnvelopes} as a regularization term to learn personalized models and the global FL model parallelly. 
Federated Mutual Learning (FML)~\cite{shen_etal} uses the non-IID nature of the data as a feature to learn personalized models.

Lastly, model interpolation techniques focus on the mixture of the local and the global models. 
In Adaptive Personalized FL (APFL)~\cite{apfl}, an optimal mixing parameter that controls the trade-off between local and global models is integrated into the learning problem. 
A recent work~\cite{mansouretal} has proposed the use of user clustering, data interpolation, and model interpolation for personalized models.
In LotteryFL~\cite{lotteryFL}, the authors adopt a Lottery Ticket Network through the application of the Lottery Ticket Hypothesis~\cite{lotteryTicketHypothesis} to learn personalized models for each user. 



\begin{table}[t!]
\centering
\caption{Example scenario to motivate the need for alternative metrics in Personalized FL.}
\resizebox{\columnwidth}{!}
{
\begin{tabular}{|c|c|c|c|c|c|c|}
\hline
\textbf{} &\textbf{Local Model} &\textbf{FedAvg} &\textbf{Per-1} &\textbf{Per-2} &\textbf{Per-3} &\textbf{Per-4} \\\cline{2-7}
\textbf{Datasplits} &\textbf{DS-1} &\textbf{DS-1} &\textbf{DS-1} &\textbf{DS-2} &\textbf{DS-3} &\textbf{DS-1} \\\cline{1-7}
\multicolumn{7}{|c|}{\textbf{Users}} \\\hline
\textbf{User0} &73\% &78\% &82\% &79\% &76\% &75\% \\ \hline
\textbf{User1} &71\% &75\% &82\% &74\% &72\% &72\% \\ \hline
\textbf{User2} &61\% &69\% &82\% &75\% &68\% &68\% \\ \hline
\textbf{User3} &55\% &71\% &75\% &79\% &82\% &96\% \\ \hline
\textbf{User4} &69\% &74\% &74\% &78\% &85\% &97\% \\ \hline
\textbf{User5} &65\% &77\% &75\% &89\% &87\% &75\% \\ \hline
\textbf{User6} &74\% &80\% &77\% &74\% &78\% &76\% \\ \hline
\textbf{User7} &68\% &82\% &77\% &76\% &79\% &78\% \\ \hline
\textbf{User8} &75\% &85\% &78\% &79\% &79\% &77\% \\ \hline
\textbf{Avg. Acc.}                  & \textbf{67.89\%} & \textbf{76.78\%} & \textbf{78\%}    & \textbf{78.11\%} & \textbf{78.44\%} & \textbf{79.33\%} \\ \hline
\end{tabular}
}
\label{table:problemStatementMotivation}
\end{table}


\begin{table*}[t!]
\centering
\setlength\tabcolsep{6pt} 
\def\arraystretch{1} 
\caption{The analysis results of studied related personalized FL methods.}
\resizebox{1\textwidth}{!}{
\begin{threeparttable}
\begin{tabular}{|c|c|c|c|c|c|c|}
\hline
\textbf{\#} & 
\textbf{Method}                                                                                 & \textbf{Only FL metrics$^\dagger$}             & \textbf{Fairness Analysis}                                                                                                        & \textbf{Datasets}$^\ddagger$ & 
\begin{tabular}[c]{@{}c@{}}
    \textbf{Use of Custom} \\ \textbf{Datasplit}$^\ast$ \\
\end{tabular} 
                                & \textbf{Comparison}      \\ \hline
%
\rowcolor{Gray}
\multicolumn{7}{|c|}{\textbf{\texttt{LOCAL FINE-TUNING}}}\\\hline
$\mathtt{1}$ & \textbf{APFL}~\cite{apfl}                                                                                     & \checkmark                                                                                                                & \xmark                                                   & \begin{tabular}[c]{@{}c@{}}  (1), (2), (3), (6) \end{tabular}                   & \checkmark                                                                                                                                        & \begin{tabular}[c]{@{}c@{}}FedAvg, SCAFFOLD,\\ Per-FedAvg, pFedMe\end{tabular} \\ \hline

$\mathtt{2}$ & \textbf{pFedMe}~\cite{pFedMe}                                                                                   & \checkmark                                                                                                               & \xmark                                                                                                               & (1), (6)                      & \checkmark                                                                                                                                        & FedAvg, Per-FedAvg                                                             \\ \hline

$\mathtt{3}$ & \textbf{Per-FedAvg}~\cite{metaFL_2}                                                                               & \checkmark                                                                                                                & \xmark                                                                                                          & (1), (3)                & \checkmark                                                                                                                                               & FedAvg                                                                         \\ \hline

$\mathtt{4}$ & \textbf{FedPer}~\cite{fedPer}                                                                                   & \checkmark                                                                                                                 & \xmark                                                      & \begin{tabular}[c]{@{}c@{}} (3), (4), (5) \end{tabular}                     & \checkmark                                                                                                                                          & FedAvg                                                                         \\ \hline

$\mathtt{5}$ & \textbf{\begin{tabular}[c]{@{}c@{}} Three Approaches \\ for Personalization \end{tabular}} \cite{mansouretal}                & \checkmark                                                                                                                 &  \xmark                                                                                                            & (1), (2)                    & \checkmark                                                                                                                                        & FedAvg, AGNOSTIC                                                               \\ \hline

$\mathtt{6}$ & \textbf{\begin{tabular}[c]{@{}c@{}}Personalized FedAvg\\ \end{tabular}}~\cite{metaFL_1}             & \checkmark                                                                                                                 & \xmark                                                                                                           & (2), (7)                   & \checkmark                                                                                                                                          & FedAvg                                                              \\ \hline

$\mathtt{7}$ & \textbf{FedMeta}~\cite{fedMeta}                                                                                  & \checkmark                                                                                                                 &   \begin{tabular}[c]{@{}c@{}}
\checkmark
\end{tabular}              & \begin{tabular}[c]{@{}c@{}} (7), (8), (9), (10) \end{tabular}                   &   \checkmark                                 & FedAvg                                                                         \\ \hline
\rowcolor{Gray}
\multicolumn{7}{|c|}{\textbf{\texttt{MULTI-TASK LEARNING}}}\\\hline

$\mathtt{8}$ & \textbf{MOCHA}~\cite{fl_mtl}                                                                                    & \begin{tabular}[c]{@{}c@{}}
\checkmark
\end{tabular}                                         & \xmark           & \begin{tabular}[c]{@{}c@{}} (11), (12), (13) \end{tabular}                   &   \checkmark                                                                                                                                       & FedAvg                                                                         \\ \hline
\rowcolor{Gray}
\multicolumn{7}{|c|}{\textbf{\texttt{CONTEXTUALIZATION}}}\\\hline

$\mathtt{9}$ & \textbf{LG-FedAvg}~\cite{lgFedAvg}                                                                                & \begin{tabular}[c]{@{}c@{}}
\checkmark
\end{tabular} & \begin{tabular}[c]{@{}c@{}}
\xmark
\end{tabular} & \begin{tabular}[c]{@{}c@{}} (6), (1), (3), (14)
\end{tabular} & 
\checkmark & 
FedAvg, FEDPROX                                                                \\ \hline
\rowcolor{Gray}
\multicolumn{7}{|c|}{\textbf{\texttt{MODEL REGULARIZATION BASED PERSONALIZATION}}}\\\hline

$\mathtt{10}$ & \textbf{\begin{tabular}[c]{@{}c@{}}FedAMP\\ \end{tabular}}~\cite{fedAMP} & \checkmark                                                                                                               & \xmark                                                   & \begin{tabular}[c]{@{}c@{}} (1), (15), (2), (4) \end{tabular}                   &     \checkmark                                                                                                                                     & \begin{tabular}[c]{@{}c@{}}SCAFFOLD, APFL, \\ FedAvg, FEDPROX\end{tabular}     \\ \hline

$\mathtt{11}$ & \textbf{\begin{tabular}[c]{@{}c@{}}FML\\ \end{tabular}}~\cite{shen_etal}             & \checkmark                                                                                                                 &    \xmark                                                      & \begin{tabular}[c]{@{}c@{}} (1), (3), (4) \end{tabular}                   &     \checkmark                                                                                                                                     & FedAvg, FEDPROX                                                                \\ \hline
\rowcolor{Gray}
\multicolumn{7}{|c|}{\textbf{\texttt{MODEL INTERPOLATION BASED PERSONALIZATION}}}\\\hline

$\mathtt{12}$ & \textbf{LotteryFL}~\cite{lotteryFL}                                                                                & \checkmark                                                                                                                 & \xmark                                                             & \begin{tabular}[c]{@{}c@{}} (1), (3), (2) \end{tabular}                   & \checkmark                                                                                                                                         & FedAvg, LG-FedAvg                                                              \\ \hline
\end{tabular}
$\dagger$ Whether the personalization method only reports metrics in the \texttt{FL} domain such as training loss, average validation accuracy, prediction error, and the number of communication rounds.
$\ddagger$ (1) MNIST, (2) EMNIST, (3) CIFAR-10, (4) CIFAR-100, (5) FLICKR-AES, (6) Synthetic, (7) Shakespeare, (8) FEMNIST, (9) Sentiment 140, (10) Industrial recommendation task, (11) Google Glass (GLEAM), (12) Human Activity Recognition (HAR), (13) Vehicle Sensor, (14) Mobile Assessment for Prediction of Suicide (MAPS), (15) FMNIST.
$\ast$ Whether the personalization method uses a custom data split technique.
\end{threeparttable}
}
\label{table:relatedWorks}
\end{table*}

\section{Problem Statement}
\label{sec:problemStatement}
We have studied $12$ recent personalized FL methods to identify their datasets, data-split strategies, the metrics other than those commonly used in the FL settings, the approaches they are compared to, and whether they perform a fairness analysis (See Table~\ref{table:relatedWorks}).
We observed two main issues in their evaluation of personalized models, which hinder the interpretability of the personalized FL methods. 
Below we provide an example scenario and present these issues.

\shortsectionBf{Motivating Example.}
We consider $\mathtt{9}$ users that collaboratively learn a global model for the next-character prediction on the keypad of their mobile phones. 
The dataset is distributed to each user to mimic the non-IID nature of real-world data distributions based on a particular data-split strategy (\textbf{DS-1}, \textbf{DS-2}, and \textbf{DS-3}).
Each user learns a local model, a global model (\texttt{FedAvg}), and a personalized model using four different personalized FL methods (\textbf{Per-1}-\textbf{Per-4}).
The personalized models are specific to each user and aim to yield better accuracy than local and \texttt{FedAvg} models.
Table~\ref{table:problemStatementMotivation} presents the accuracy of local models, \texttt{FedAvg} model, and four personalized models on different datasplits.

\shortsectionBf{Missing Per-User Accuracy and Fairness Analysis.} 
The personalized FL methods often solely report the average accuracy of the personalized model across all the users to measure their model effectiveness.
In Table~\ref{table:problemStatementMotivation}, we ask the question of \textit{which personalization method yields the best performance in terms of per-user personalized accuracy}.
In this example, \textbf{Per-4} gives the highest average accuracy of $\mathtt{79.3\%}$ on \textbf{DS-1}.
However, upon closer inspection, we observe that, with respect to the \texttt{FedAvg} model, \textbf{Per-4} increases the accuracy of only $2$ out of the $9$ users.
This shows that the average accuracy of the personalized model may fail to fully characterize the quality of a method.
Here another observation is that the best performing personalized method, on average, may not necessarily lead to an improvement over the local or global models across all users.

Through surveyed methods in Table~\ref{table:relatedWorks}, we observe that the methods commonly adapt the evaluation metrics from FL, such as the training loss, average validation accuracy, prediction error, number of communication rounds.
This means that the methods often do not report the per-user accuracy (Table~\ref{table:relatedWorks} ``Only FL metrics'' column).
Out of $12$ studied methods, only the FedMeta approach performs a fairness analysis among personalized models of users by reporting the per-user accuracy of their method. 

\shortsectionBf{Inconsistent Datasets and Data-splits.} 
A data-split strategy is used in personalized FL to split the dataset across users such that each receives a fraction of the non-IID data.
The data distribution among users is a crucial feature in personalization because if the data is distributed IID, the personalized model may not offer any benefits over \texttt{FedAvg}~\cite{apfl}.
The personalized methods often use the same dataset yet the data-splits are different. For instance, in Table~\ref{table:problemStatementMotivation}, \textbf{Per-1}, and \textbf{Per-4} are trained on \textbf{DS-1}, whereas \textbf{Per-2} is on \textbf{DS-2} while \textbf{Per-3} is on \textbf{DS-3}. 
The use of different data splits on the same dataset makes it difficult to interpret the effectiveness of personalized models.
When personalized models are compared with each other, each method often reports their results on a new (different) data split than the one used in the compared approach.
For example, \textbf{Per-1} and \textbf{Per-2} cannot be directly compared as they are trained on different data-splits, \textbf{DS-1} and \textbf{DS-2}.

Out of the $12$ studied personalized FL methods, we observe that $9$ methods use standard datasets such as MNIST and CIFAR-10 ((1)-(4), (6), and (7), Table~\ref{table:relatedWorks} ``Datasets column'').  
Other $3$ methods use other datasets, such as FEMNIST and Sentiment 140 ((5), and (8)-(15)).
Additionally, the methods often use different custom data splits on the datasets. 
We identified $10$ data-splits in $12$ methods (Table~\ref{table:relatedWorks} ``Use of Custom Datasplit'' column).

\section{Evaluation Metrics}
\label{sec:evaluationMetrics}
We present a set of metrics for the evaluation of personalized FL models from the performance and fairness perspective. 
%
To quantify the per-user accuracy improvements gained in terms of personalization, we compute the metrics on the Quantum of Improvement (\texttt{QoI}) as follows:
 \begin{equation}
    \mathtt{
     F_{i} = P_{i} - max(G_{i}, L_{i})
    }
    \label{eqn:qoi}
\end{equation}
where $\mathtt{P}$, $\mathtt{G}$, and $\mathtt{L}$ refer to the accuracy of the personalized model, \texttt{FedAvg} and local model of the user $\mathtt{i}$, and $\mathtt{F_{i}}$ refers to the \texttt{QoI} of user $\texttt{i}$. Henceforth, we will refer to the \texttt{QoI} as $\mathtt{F}$ in all the equations. 
%
%

The \texttt{QoI} can result in negative values. 
This means that the personalized method decreases a user's personalized model accuracy rather than the expected increase over the local or global models.  
In such cases, the direct application of evaluation metrics may misguide the interpretation of results.
Therefore, we split the \texttt{QoI} into two sets that contain the absolute \texttt{QoI} values, \ie a set of users ($\mathtt{U^{+}}$) who have positive \texttt{QoI}, and a set of users ($\mathtt{U^{-}}$) who have negative \texttt{QoI}.  
We then apply the introduced metrics below to both sets and interpret accordingly.

\subsection{Performance Metrics}
\label{subsec:performance_metrics}
We introduce five performance metrics to express how well the personalized model performs over each user's local and global model.

\shortsectionBf{Percentage of User-models Improved (PUI).} \texttt{PUI} is the percentage of users who experience an improvement over their local and global models. 
Ideally, a personalized model is expected to improve the per-user accuracy of a maximal set of users.
\begin{equation}
    \mathtt{
        PUI = \frac{COUNT(F_{i} > 0)}{COUNT(U)} \times 100, i \in U (U: Users)
    }
\end{equation}

%






In a normal distribution, the mean is the best measure of the central tendency. 
However, the median might be a better measure of central tendency when this is not the case. 
We define median and average percentage of improvement since the \texttt{QoI} distribution is not known apriori. 

\shortsectionBf{Median Percentage of Improvement (MPI).} \texttt{MPI} is computed as $\mathtt{Median(U^{+})}$ where $\mathtt{Median()}$ function returns the median of its input, and $\mathtt{U^{+}}$ is the \texttt{QoI} of the set of users who obtained an increase in their performance.
A personalized model is expected to have a high median of the \texttt{QoI} values among the users who experience an improvement.

\shortsectionBf{Average Percentage of Improvement (API).} \texttt{API} is the average percentage improvement among the users who obtained an increase in their performance ($\mathtt{U^{+}}$). 
\begin{equation}
    \mathtt{
        API = \frac{\sum_{i \in U^{+}} F_{i}}{len(U^{+})}
    }
\end{equation}



We observe that, in some scenarios, a personalization method does not yield an improvement over users' local and global accuracy. Thus,in such cases, it is crucial to report the per-user accuracy decrease of the personalized model. 
Because this decrease cannot be derived from the improvement metrics (MPI and API), we define two metrics to quantify the decreased accuracy. 
%


%






\shortsectionBf{Median Percentage of Decrease (MPD).}
Similar to \texttt{MPI}, \texttt{MPD} is computed as $\mathtt{Median(U^{-})}$. 


\shortsectionBf{Average Percentage of Decrease (APD).} Similar to \texttt{API}, \texttt{APD} is the average percentage decrease among the users whose performance is decreased ($\mathtt{U^{-}}$).


\subsection{Fairness Metrics}
\label{subsec:fairness_metrics}
We extend four metrics to evaluate personalization methods that yield better results from a fairness perspective.
For two personalization methods $\mathtt{t}$ and $\mathtt{t'}$, the \texttt{QoI} distribution 
among $\mathtt{K}$ users $\mathtt{\{F_{1}(t), \ldots, F_{K}(t)\}}$ is more fair (uniform) under technique $\mathtt{t}$ than $\mathtt{t'}$ based on the relation captured by the fairness metric.

\shortsectionBf{Average Variance (AV).} 
Average Variance (\texttt{AV})~\cite{qFFL} is a measure of the spread of data.
For \texttt{AV}, the relation is extended to personalized models as follows:
\begin{equation}
    \textbf{$\mathtt{AV}$}\mathtt{(F_{1}(t), ... F_{K}(t))} < \textbf{$\mathtt{AV}$}\mathtt{(F_{1}(t'), ... F_{K}(t'))}
\end{equation}
where, $\mathtt{AV(F_{1}(t), ... F_{K}(t))}$ is computed as,
\begin{equation}
    \mathtt{
        AV = \frac{1}{K} \sum^{K}_{i=1} (F_{i}(t) - \Bar{F}(t))^{2}
    }
    \label{eqn:Var}
\end{equation}





$\mathtt{\Bar{F}(t)}$ in Equation~\ref{eqn:Var} refers to the average \texttt{QoI} across all users and is computed as, 
\begin{equation}
\mathtt{\Bar{F}(t) = \frac{1}{K} \sum^{K}_{i=1} (F_i(t))}
\end{equation}
A lower \texttt{AV} means a higher fairness capability for a personalized method. 


\shortsectionBf{Cosine Similarity (CS).} 
One of the drawbacks of \texttt{AV} is that the outliers may cause skewing of the data.
We measure Cosine Similarity (\texttt{CS})~\cite{qFFL} for two personalized methods to quantify the similarity between their \texttt{QoI} distributions. For \texttt{CS}, the relation is computed as:

\begin{equation}
    \textbf{$\mathtt{CS}$}\mathtt{ [(F_{1}(t), ... F_{K}(t)), 1]} \geq \textbf{$\mathtt{CS}$}\mathtt{ [(F_{1}(t'), ... F_{K}(t')), 1]}
\end{equation}
where, $\mathtt{ CS[(F_{1}(t), ... F_{K}(t)), 1] }$ is computed as follows.
%
\begin{equation}
    \mathtt{
        CS = \frac{\frac{1}{K} \sum^{K}_{i=1} F_{i}(t)}{\sqrt{\frac{1}{K} \sum^{K}_{i=1} F^{2}_{i}(t)}}
    }
\end{equation}
A higher \texttt{CS} means a higher fairness capability for a personalized method.

\shortsectionBf{Entropy.} 
One of the drawbacks of \texttt{CS} is that the magnitude of the QoI values is not taken into consideration, yet only their orientation is considered.
For this reason, we extend \texttt{Entropy}~\cite{qFFL} as follows:
\begin{equation}
    \textbf{$\mathtt{Entropy}$}\mathtt{(F_{1}(t), ... F_{K}(t))} \geq \textbf{$\mathtt{Entropy}$}\mathtt{(F_{1}(t'), ... F_{K}(t'))}
\end{equation}
where $\mathtt{Entropy}$ is defined as follows.
\begin{equation}
    \mathtt{
        Entropy = - \sum^{K}_{i=1} \frac{F_{i}(t)}{\sum^{K}_{i=1}F_{i}(t)}log(\frac{F_{i}(t)}{\sum^{K}_{i=1}F_{i}(t)})
    }
\end{equation}
A higher \texttt{Entropy} means a higher fairness capability for a personalized method.

\shortsectionBf{Jain's index (JI).} \texttt{JI} is widely studied fairness measure in computer networks and resource allocation to identify the underutilized channels~\cite{jains_index}. 
We extend it for personalization as follows:
\begin{equation}
    \mathtt{
     JI = \frac{[\sum^{K}_{i=1}F_{i}(t)]^{2}}{K \sum^{K}_{i=1}F_{i}(t)^2}
    }
\end{equation}
%
A higher \texttt{JI} means a higher fairness capability for a personalized method.
\section{Experimental Results}
\label{sec:evaluations}

\begin{table*}[th!]\centering
\def\arraystretch{1.0}
\setlength\tabcolsep{8pt}
\caption{Per-user accuracy for the different personalized FL methods on the different data splits of the CIFAR-10 dataset.}
\resizebox{\textwidth}{!}
{
\begin{tabular}{l|rrr|rrr|rrr|rrr|rrrr}\toprule
\textbf{} &\multicolumn{3}{c|}{\textbf{FedAvg}} &\multicolumn{3}{c|}{\textbf{PersFL}} &\multicolumn{3}{c|}{\textbf{FedPer}} &\multicolumn{3}{c|}{\textbf{pFedMe}} &\multicolumn{3}{c}{\textbf{Per-FedAvg}} \\
\cmidrule{1-16}
\textbf{Users} &\textbf{DS-1} &\textbf{DS-2} &\textbf{DS-3} &\textbf{DS-1} &\textbf{DS-2} &\textbf{DS-3} &\textbf{DS-1} &\textbf{DS-2} &\textbf{DS-3} &\textbf{DS-1} &\textbf{DS-2} &\textbf{DS-3} &\textbf{DS-1} &\textbf{DS-2} &\textbf{DS-3} \\
\midrule
\rowcolor{Gray}
\textbf{User 0} &43.6 &50.8 &48.2 &85.5 &61.3 &94.5 &83.2 &57.2 &93.1 &74.3 &61.7 &94.2 &69.2 &58.2 &92.5 \\
\textbf{User 1} &50.9 &45.3 &40.8 &78.2 &56.9 &79.9 &74.5 &51.4 &77 &64.1 &57.3 &79.2 &65 &56.1 &73.7 \\
\rowcolor{Gray}
\textbf{User 2} &44.5 &49.4 &31.2 &82.2 &57.3 &68.9 &78.4 &53.2 &64.7 &69.6 &57.3 &64.4 &67.9 &57.2 &64.6 \\
\textbf{User 3} &51.3 &46.5 &31.5 &82.1 &60.1 &82.5 &77.9 &55.4 &77.5 &69.4 &58.9 &72.5 &67.2 &58.8 &77 \\
\rowcolor{Gray}
\textbf{User 4} &45.3 &50.8 &49.4 &79.4 &59.1 &82.5 &76.1 &54.4 &78.6 &67.2 &59.5 &80 &65.8 &59.4 &82.5 \\
\textbf{User 5} &44.2 &50.7 &47.8 &77.1 &61.9 &79.9 &72.1 &57.6 &76.9 &62.1 &60.6 &77.5 &62.7 &59.3 &77.9 \\
\rowcolor{Gray}
\textbf{User 6} &35.8 &46.5 &56.8 &75.6 &58.9 &90.3 &70.9 &53.3 &88.5 &59.9 &58.6 &88.3 &58.2 &57.7 &89.1 \\
\textbf{User 7} &37.9 &49.5 &58.1 &79.7 &61.2 &87.6 &75.6 &56.9 &84.6 &65.8 &60.2 &84 &64.3 &58 &83.7 \\
\rowcolor{Gray}
\textbf{User 8} &47.7 &48.7 &49 &87.7 &60 &76.7 &84.4 &57.5 &73.5 &75.5 &59.6 &66.9 &72.5 &55.8 &64.1 \\
\textbf{User 9} &48.6 &49.1 &53.5 &91 &58.8 &80.3 &88.5 &54.3 &77.9 &81.7 &58.3 &73.8 &76.6 &55.3 &72.7 \\
\midrule
\textbf{Avg. Acc.} &\textbf{45} &\textbf{48.7} &\textbf{46.6} &\textbf{81.9} &\textbf{59.6} &\textbf{82.3} &\textbf{78.2} &\textbf{55.1} &\textbf{79.2} &\textbf{69} &\textbf{59.2} &\textbf{78.1} &\textbf{66.9} &\textbf{57.6} &\textbf{77.8} \\
\midrule
\textbf{Std Dev} &\textbf{5.1} &\textbf{2} &\textbf{9.4} &\textbf{4.9} &\textbf{1.7} &\textbf{7.2} &\textbf{5.6} &\textbf{2.1} &\textbf{7.9} &\textbf{6.7} &\textbf{1.4} &\textbf{9.2} &\textbf{5.1} &\textbf{1.5} &\textbf{9.5} \\
\bottomrule
\end{tabular}
}
\label{tab:Results_CIFAR10}
\end{table*}

\begin{table*}[t!]
\centering
\caption{The performance metrics applied to the QoI of different personalized FL methods on CIFAR-10 dataset.}
\setlength\tabcolsep{10pt} 
\def\arraystretch{1} 
\resizebox{\textwidth}{!}
{
\begin{threeparttable}
\begin{tabular}{l|rrr|rrr|rrr|rrr|r}
\toprule
\textbf{} &\multicolumn{3}{c|}{\textbf{PersFL}} &\multicolumn{3}{c|}{\textbf{FedPer}} &\multicolumn{3}{c|}{\textbf{pFedMe}} &\multicolumn{3}{c|}{\textbf{PerFed}} \\
\cmidrule{2-13}
\textbf{Metrics} &\textbf{DS-1} &\textbf{DS-2} &\textbf{DS-3} &\textbf{DS-1} &\textbf{DS-2} &\textbf{DS-3} &\textbf{DS-1} &\textbf{DS-2} &\textbf{DS-3} &\textbf{DS-1} &\textbf{DS-2} &\textbf{DS-3} \\
\midrule
\rowcolor{Gray}
\begin{tabular}[c]{@{}c@{}} \texttt{PUI} \end{tabular} 
&100 &100 &100 &100 &100 &100 &100 &100 &100 &100 &100 &100 \\
\begin{tabular}[c]{@{}c@{}} \texttt{MPI} \end{tabular}
&38.74 &11.23 &33.29 &34.53 &6.59 &30.43 &24.58 &10.81 &31.06 &22.9 &8.55 &32.65 \\
\rowcolor{Gray}
\begin{tabular}[c]{@{}c@{}} \texttt{API} \end{tabular}
&36.87 &10.83 &35.67 &33.18 &6.41 &32.59 &23.98 &10.47 &31.44 &21.95 &8.85 &31.17 \\
\begin{tabular}[c]{@{}c@{}} $\texttt{AA}^{\dagger}$ \end{tabular}
&81.85 &59.56 &82.29 &78.16 &55.13 &79.21 &68.95 &59.19 &78.07 &66.93 &57.57 &77.79 \\
\bottomrule
\end{tabular}
{\small{$\texttt{AA}^{\dagger}$ is the average accuracy across all users.}}
\end{threeparttable}
}
\label{tab:performance_CIFAR10}
\end{table*}

We evaluate four recently introduced personalization methods, \sys~\cite{persFL}, \texttt{FedPer}~\cite{fedPer}, \texttt{pFedMe}~\cite{pFedMe}, and \texttt{Per-FedAvg}~\cite{metaFL_2}, to evaluate their performance and fairness through introduced metrics.

\subsection{Experimental Setting} 
We use CIFAR-10 dataset on three data splits with a total of $\mathtt{10}$ clients. CIFAR-10 includes $\mathtt{32\times32}$ color images with $\mathtt{10}$ classes and $\mathtt{60,000}$ instances. 
We use a CNN-based model with two 2-D convolutional layers separated by a MaxPool layer between them and followed by three fully connected (FC) layers.  
The fully connected layers have $400$, $120$, and $84$ hidden neurons. We use ReLu activations after each layer except the last FC layer.

We make three assumptions in line with the assumptions made in personalized FL literature. 
First, we assume that all clients are active during the entire training phase to speed up the model convergence. 
Second, each client's data does not change between the global aggregations. 
Lastly, the hyper-parameters, batch-size ($\mathtt{B}$), and local epochs ($\mathtt{E}$), are invariant among the simulated clients.
We conduct all experiments with a $\mathtt{60\%}$-$\mathtt{20\%}$-$\mathtt{20\%}$ train-validation-test splits.

The experiments are run with Python $\mathtt{3.7}$, and a PyTorch version of $\mathtt{v1.3.1}$ on an NVIDIA Tesla T4 GPU with $\mathtt{16GB}$ memory with a CUDA version of $10.0$. 
%

\shortsectionBf{Data-splitting Strategies.}
We split the data among users following three different strategies used in the literature.

In \textbf{DS-1}, each user has the same total number of samples but may have different classes and a different number of samples per class.
The statistical heterogeneity is varied by tuning the parameter $\mathtt{k}$, which controls the number of overlapping classes between each user~\cite{fedPer}.
%
For example, $\mathtt{k=4}$ corresponds to a highly non-identical data partition, whereas $\mathtt{k=10}$ corresponds to a highly identical data partition across the participating users. We set $\mathtt{k}$ to 4 to have non-IID data across users. 

In \textbf{DS-2}~\cite{salvage_localAdaptation}, all users have samples from all classes, but the number of samples per class they have is different, and hence the total number of samples per user is also different across users. 
In order to simulate a non-IID distribution, we assign samples from each class to the users using a Dirichlet distribution with $\mathtt{\alpha = 0.9}$, following the previous work~\cite{nIID_dist}. Each class is parameterized by a vector $\mathtt{q}$ where $\mathtt{q \geq 0, i \in [1, N]}$ 
is sampled from a Dirichlet distribution with parameters $\mathtt{\alpha}$ and $\mathtt{p}$. 
The parameter $\mathtt{p}$ is the prior class distribution over the classes, and $\mathtt{\alpha}$ is the concentration parameter that controls the data similarity among the users. 
If $\mathtt{\alpha \rightarrow \infty}$, all users have an identical distribution to the prior. If $\mathtt{\alpha \rightarrow 0}$, each user only has samples from one class randomly chosen. 

In \textbf{DS-3}, each user has two of the ten class labels. 
Additionally, the total number of samples per user is different, \ie the users do not have the same number of total samples. 
The samples assigned to users are drawn from a log-normal distribution with the parameters $\mathtt{\mu = 0}$ and $\mathtt{\sigma = 2}$~\cite{pFedMe}.
%
%
The parameters $\mathtt{\mu}$ and $\mathtt{\sigma}$ correspond to the underlying normal distribution from which we draw samples.

\subsection{Experimental Results}
\label{subsec:experimentalResults}
We evaluate four personalized learning methods to answer the following two questions: 

\begin{enumerate}[itemsep=0mm]
\item Which personalization method performs the best in terms of per-user personalized accuracy across all users? 
\item Which algorithm is the fairest?
\end{enumerate}

Table~\ref{tab:Results_CIFAR10} presents the per-user accuracy of personalized models and \texttt{Fed-Avg} across different data splits of CIFAR-10.

\begin{figure*}[th!]
    \centering 
    \subfigure[\textbf{DS-1}]{\includegraphics[width=0.33\textwidth]{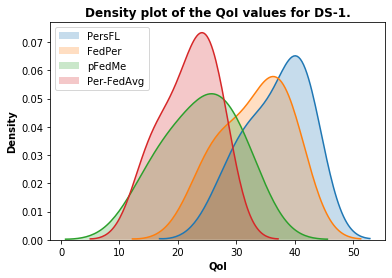}} 
    \subfigure[\textbf{DS-2}]{\includegraphics[width=0.33\textwidth]{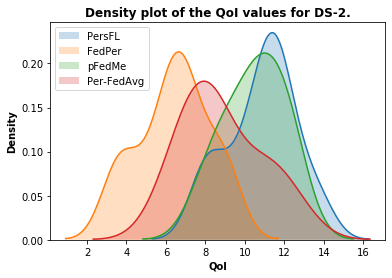}} 
    \subfigure[\textbf{DS-3}]{\includegraphics[width=0.33\textwidth]{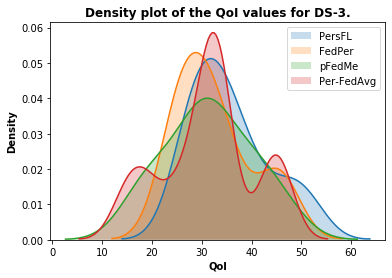}}
    \caption{Density plots of \texttt{QoI} across all data splits of CIFAR-10 for each personalized FL method.}
    \label{fig:densityPlots}
\end{figure*}

\subsubsection{Performance Metrics}
\label{subsubsec:performancePerspective}
We use the average accuracy (\texttt{avg-acc}) in conjunction with introduced metrics to make a more informed decision on evaluating personalized approaches instead of solely using \texttt{avg-acc}.
Table~\ref{tab:performance_CIFAR10} shows the performance metrics applied to the \texttt{QoI} of the different personalization methods across different data splits of CIFAR-10.
We report a subset of the performance metrics (\texttt{PUI}, \texttt{MPI}, and \texttt{API}) since all the personalization methods that we have surveyed lead to an increase in the personalized per-user accuracy in our experiments. 
%
%
This means that the \texttt{PUI} for each personalization method yields $\mathtt{100\%}$ with none of the users experiencing a decrease over their local and global models.
%

In terms of \texttt{MPI} and \texttt{API} on \textbf{DS-1}, \sys performs the best at $\mathtt{38.74\%}$ and $\mathtt{36.87\%}$.
On \textbf{DS-2}, \sys and \texttt{pFedMe} yield the highest \texttt{MPI} at $\mathtt{11.23\%}$ and $\mathtt{10.81\%}$. 
Similarly, these methods lead to an \texttt{API} of $\mathtt{10.83\%}$ and $\mathtt{10.47\%}$. 
We observe that on \textbf{DS-3}, \sys has the highest \texttt{MPI} and \texttt{API}, at $\mathtt{33.29\%}$ and $\mathtt{35.67\%}$. 
In terms of \texttt{avg-acc},  \sys achieves the highest accuracy $\mathtt{81.85\%}$ on \textbf{DS-1}. 
On \textbf{DS-2}, \sys and \texttt{pFedMe} are the best performing at $\mathtt{59.56\%}$, and $\mathtt{59.19\%}$. Lastly, on \textbf{DS-3}, \sys is the top-performing method at $\mathtt{82.29\%}$.
%


%

\begin{figure*}[th!]
\begin{center}
\centerline{\includegraphics[width=2.1\columnwidth]{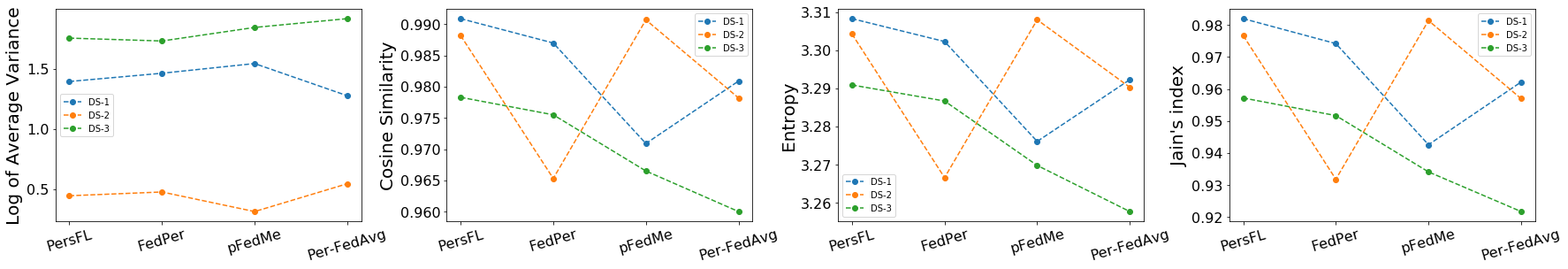}}
\caption{Fairness metrics applied to the personalization methods across all data splits of the CIFAR-10 dataset.}
\label{fig:fairness_CIFAR10}
\end{center}
\end{figure*}

Figure~\ref{fig:densityPlots} shows the density plots of the \texttt{QoI} for all users across data splits of the CIFAR-10 dataset. 
The peaks in a density plot show the values concentrated over an interval where the x-axes of the plots show the \texttt{QoI} intervals and the y-axes show the density.
We observe on \textbf{DS-1} that the most values for \sys are concentrated around a higher value in the range of \texttt{QoI} values.
On \textbf{DS-2}, the peak of \sys is associated with a higher \texttt{QoI} interval compared to \texttt{pFedMe}.
However, the distribution of \texttt{pFedMe} is more normal compared to \sys as \sys has an additional peak corresponding to the \texttt{QoI} interval of $\mathtt{8}$.
On \textbf{DS-3}, the \texttt{QoI} interval corresponding to the peaks of \sys, \texttt{pFedMe}, and \texttt{Per-FedAvg} gives almost the same \texttt{QoI} value, where their \texttt{QoI} is concentrated.

\subsubsection{Fairness Metrics}
\label{subsubsec:fairnessPerspective}
We apply fairness metrics across all the data splits of CIFAR-10, as shown in Figure~\ref{fig:fairness_CIFAR10}.
The x-axis represents the different personalization methods, and the y-axis is the different fairness metrics (\texttt{AV}, \texttt{CS}, \texttt{Entropy}, and \texttt{JI}) applied to these methods. 
%
We observe in Figure~\ref{fig:fairness_CIFAR10} that the trends in metrics are similar across the different data splits. 
The main reason is that \texttt{QoI} distribution of personalized methods for the CIFAR-10 dataset is identical to each other. However, different trends can be observed with the application of methods to different datasets.
%
We note that for a method to be fair, 
it also needs to give a reasonable performance in terms of the per-user personalization accuracy.

In Figure~\ref{fig:fairness_CIFAR10}, \texttt{Per-FedAvg} gives the lowest \texttt{AV} amongst all methods.
A lower \texttt{AV} means better per-user personalization accuracy distribution as the method yields more uniform accuracy than the other algorithms. 
For other three metrics (\texttt{CS}, \texttt{Entropy}, and \texttt{JI}), \sys performs the best. 
The higher the value of these metrics, the fairer \texttt{QoI} distribution. 
Therefore, on \textbf{DS-1}, \sys is the fairest amongst the evaluated personalization methods.


On \textbf{DS-2}, \texttt{pFedMe} has the least \texttt{AV}, and the highest value for the other three metrics.
This confirms that it generalizes per-user personalized accuracy across all the users. 
%
On \textbf{DS-2} in Table~\ref{tab:performance_CIFAR10}, 
the per-user personalization accuracy of \texttt{pFedMe} is similar to that of \sys from a performance perspective. At the same time, \texttt{pFedMe} is relatively fairer than \sys, therefore \texttt{pFedMe} is the fairest algorithm on \textbf{DS-2}.

Lastly, on \textbf{DS-3}, from Figure~\ref{fig:fairness_CIFAR10},  \texttt{FedPer} has the lowest \texttt{AV}. 
However, it is not the fairest algorithm according to the other fairness metrics.
Among the other three metrics, \sys gives the highest values.

\section{Conclusions}
\label{sec:conclusions}
We introduce and adapted new metrics for performance and fairness, complementing the widely reported average personalized model accuracy to evaluate the personalization methods.
We employed these metrics on four recent personalized FL methods across three different data splits on the CIFAR-10 dataset.
Our evaluation results show that the personalized model that gives the highest average accuracy across users is not necessarily the fairest.

\bibliography{paper}
\bibliographystyle{icml2021}





\end{document}